\documentclass[10pt,twocolumn,letterpaper]{article}
\pdfoutput=1
\usepackage{iccv}
\usepackage{times}
\usepackage{epsfig}
\usepackage{graphicx}
\usepackage{amsmath}
\usepackage{amssymb}
\usepackage{algorithmicx}
\usepackage{algpseudocode}
\usepackage{amsmath,macro}
\usepackage{multirow}
\usepackage{floatrow}
\makeatletter
\newcommand*{\rom}[1]{\expandafter\@slowromancap\romannumeral #1@}

\usepackage[breaklinks=true,bookmarks=false]{hyperref}

\iccvfinalcopy 

\def\iccvPaperID{****} 

\ificcvfinal\fi

\begin{document}

\title{Poisoning MorphNet for Clean-Label Backdoor Attack to Point Clouds}

\author{Guiyu Tian\\
Peking University \\
{\tt\small guiyutian@pku.edu.cn}
\and
Wenhao Jiang\\
Tencent\\
{\tt\small cswhjiang@gmail.com}
\and
Wei Liu \\
Tencent\\
{\tt\small wl2223@columbia.edu}
\and
Yadong Mu\\
Peking University \\
{\tt\small myd@pku.edu.cn}
}

\maketitle
\ificcvfinal\thispagestyle{empty}\fi
\begin{abstract}



This paper presents Poisoning MorphNet, the first backdoor attack method on point clouds. Conventional adversarial attack~\cite{adt_survey} takes place in the inference stage, often fooling a model by perturbing samples. In contrast, backdoor attack~\cite{BadNet} aims to implant triggers into a model during the training stage, such that the victim model acts normally on the clean data unless a trigger is present in a sample. This work follows a typical setting of clean-label backdoor attack, where a few poisoned samples (with their content tampered yet labels unchanged) are injected into the training set. 
The unique contributions of MorphNet are two-fold. First, it is key to ensure the implanted triggers both visually imperceptible to humans and lead to high attack success rate on the point clouds. To this end, MorphNet jointly optimizes two objectives for sample-adaptive poisoning: a reconstruction loss that preserves the visual similarity between benign / poisoned point clouds, and a classification loss that enforces a modern recognition model of point clouds tends to mis-classify the poisoned sample to a pre-specified target category. This implicitly conducts spectral separation over point clouds, hiding sample-adaptive triggers in fine-grained high-frequency details. Secondly, existing backdoor attack methods are mainly designed for image data, easily defended by some point cloud specific operations (such as denoising). We propose a third loss in MorphNet for suppressing isolated points, leading to improved resistance to denoising-based defense. Comprehensive evaluations are conducted on ModelNet40 and ShapeNetcorev2. Our proposed Poisoning MorphNet outstrips all previous methods with clear margins.

\end{abstract}

\section{Introduction}

Deep neural networks have re-calibrated the performance records for a large spectrum tasks that tackle image~\cite{resnet}, video~\cite{i3d} or point cloud~\cite{MVCNN,voxnet}. However, deep models are known to be vulnerable to adversarial attacks~\cite{adt_survey,CW,FGSM}, which brings high risks to the practical deployment of deep models. Recently, a new attack paradigm named backdoor attack~\cite{survey1,survey2,survey3} has attracted increasing attention. Conceptually, conventional adversarial attack methods mislead the inference of a well-trained model by adding perturbation into a clean sample. On the contrary, backdoor attack assumes the access of training data. It attempts to poison a portion of data and inject them into the training set. The poisoned data contain pre-defined triggers. The main purpose of backdoor attack is to attract deep models to mis-categorize a sample with trigger, regardless of its original content. During inference, the victim model acts normally on the clean data unless a trigger is present. The hidden backdoors can pose a huge threat to the security of point cloud oriented deep models. Owing to be data-hungry, many models are trained including gigantic crowed-sourced point-clouds as in the case of learning high-definition maps from many anonymous autonomous vehicles, which leaves space for poisoned data injection.

The essence of backdoor attack is that triggers can provide a \emph{shortcut}, through which the deep model quickly makes high-confidence decision by fully skipping the sample's content. A good trigger should strike balance between stealthy trigger embedding and attack effectiveness. In this paper, we study the backdoor attack on point clouds in a clean-label setting~\cite{cleanlabel_turner}. Importantly, a poisoned sample indeed owns two labels: a ground-truth label that is consistent with its content, and an arbitrary label that trigger defines (often different from the ground-truth label). When adding a poisoned sample into the training set, it is crucial to determine which label is used during training the model. In a clean-label setting, a sample will not have label altered after being poisoned. This ensures the consistency between the content / label and thus reduces the risk of being detected by human eyes, at the cost of increasing difficulty of implanting triggers into the learned model.


Our proposed model, dubbed as Poisoning MorphNet, is strongly motivated by two considerations: 

First, existing state-of-the-art methods of backdoor attack (such as~\cite{clean_fudan,ReflectionECCV}) are mostly designed for images or videos, leaving the study on point clouds untouched. It is seemingly natural to directly borrow insights in the image domain for attacking point clouds. Nonetheless, most key insights are either image-specific (\eg,~\cite{ReflectionECCV} harnesses the low visibility of image reflection that is not applicable for point clouds), or lead to inferior performance when directly tailed to the point clouds, as revealed by our experiments. In addition, unlike images, we observe that backdoor attack models are surprisingly vulnerable to some point cloud oriented defenses, such as point cloud denoising~\cite{SOR}. This spurs more domain-specific neural architectural designs in our proposed Poisoning MorphNet for improving the model's resistance to defenses.

Secondly, we argue that sample-adaptive backdoor attack is still inadequately explored in current literature. To make the victim model more easily recognize the trigger, existing backdoor attack methods mostly learn triggers that are universal for all samples from the same category. As a main weakness under such a treatment, poisoned samples can be easily filtered out by a naive PCA (Principal Component Analysis) process~\cite{spectral_defense}. Additionally, if the attack targets multiple classes, one needs to manually design unique triggers for each class or iterate complicated class-wise optimization~\cite{clean_fudan}. Instead, our MorphNet reads both a benign sample and a target trigger label as its input and returns a poisoned sample. This renders a sample-adaptive attack, rather than on a fixed-trigger basis. This bears advantages in several-folds: poisoning that adapts to sample can arguably make the trigger less detectable by human inspectors. This is a key metric for evaluating the success of clean-label backdoor attacks. In addition, treating trigger labels as part of the model's input also makes the backdoor injection more flexible, since one can trivially generate more poisoned samples by changing the conditional class in MorphNet.




To our best knowledge, this is the first backdoor attack method for 3D point cloud. Comprehensive evaluations are conducted on ModelNet40 and ShapeNetcorev2. In all experiments our proposed Poisoning MorphNet outstrips all computing methods (mainly adapted from image-oriented methods~\cite{clean_fudan,cleanlabel_turner}) with clear and significant margins. 





\section{Related Work}

\textbf{Backdoor Attack}. It is a novel attack paradigm that is still inadequately researched. A few attack settings have been explored towards improved flexibility and robustness, including 1) \emph{clean-label attack}~\cite{cleanlabel_turner}: It has been proposed in order to better conceal the attack intention, where the poisoned sample must be visually consistent with the ground-truth label. Recent works~\cite{ReflectionECCV, clean_fudan, AAAI_hidden} under such settings either add adversarial perturbation on benign sample in advance, or design more powerful triggers. 2) \emph{sample-adaptive}: Most existing methods utilize a fixed trigger, which heavily affects the principal components of the data and can be detected by simple targeted defense method~\cite{spectral_defense}. More recent works proposed input-aware triggers~\cite{NIPS_aware} where the trigger varies between different samples. Such input-adaptive trigger is less likely to be detected by defense methods. 3) \emph{multi-label attack}: The vast majority of attack methods focus on single label attack, where only one class is selected as the target class. In this work, we consider a more difficult setting, \ie, multi label attack, where the attackers can freely attack all classes according to their will. In other words, when attacking, the attacker can generate poison samples to attack different target classes by processing the same clean sample differently. This attack setting gives the attacker greater authority and also brings greater danger to model security.

\textbf{Backdoor Defense}. Many defense methods against backdoor attacks have also been proposed. Data augmentation is a concise and general method, previously also widely-used to defend adversarial attack. Besides, Tran et al.~\cite{spectral_defense} tried to screen poisoned samples in a pre-processing step. The key insight is that, for clean-label backdoor attack, fixed triggers will bring a salient principal component in the feature space. One can thereby find data with triggers by investigating the similarity between data and the principal component. Wang et al.~\cite{neural_clean} devised \emph{Neural Cleanse} to detect whether the model has concealed a trigger. It is assumed that the perturbation required for an infected model to misclassify samples into target class is usually smaller than other classes. Thus, one can use robust statistics to detect the outliers and reveal possible trigger.

\textbf{3D Point Cloud Recognition}. 3D point clouds have diverse applications in many fields, including autonomous driving~\cite{drive}, robotics, scene reconstruction, etc. Numerous deep networks for point clouds have been built. In the early works, MVCNN \cite{MVCNN} is proposed to max-pool multi-view features into a global representation. Several following works~\cite{MV2,MV3} further improved the performance of multi-view based method. The work in~\cite{voxnet} voxelizes the point clouds into 3D grids and applies 3D CNN to obtain the final representation. Point-based methods~\cite{pointnet,pointnet++,DGCNN} empirically achieve higher performance and better computational efficiency. In this work, we mainly focus on attacking the point-based models, such as PointNet~\cite{pointnet}, DGCNN~\cite{DGCNN} and PointNet++~\cite{pointnet++}.

\section{Our Approach}

\begin{figure*}[t]
\centering
\includegraphics[width = 0.9\linewidth]{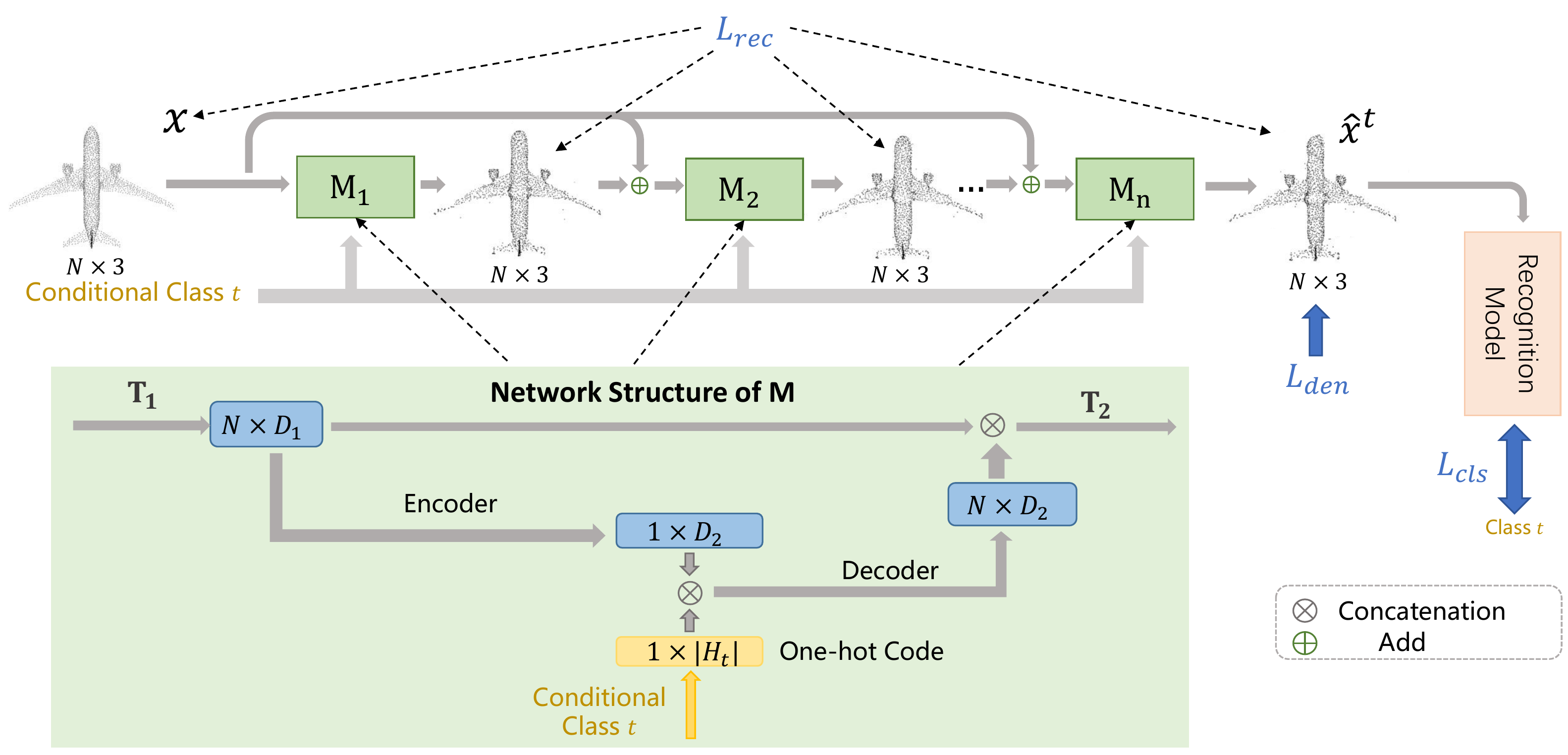}
\caption{\small Network architecture of our proposed Poisoning MorphNet.}
\label{figure:structure}
\vspace{-0.15in}
\end{figure*}


\subsection{Design of Poisoning MorphNet}

The process of generating poisoned samples in existing methods (\eg, ~\cite{clean_fudan,ReflectionECCV})) can be summarized as:
\begin{equation}
\hat{x}_i =  m \odot x_i + (1-m) \odot s,
\end{equation}
where $x_i, \hat{x}_i$ denote a clean sample and its poisoned version, respectively. $m$ is a binary mask and $s$ represents the trigger. $\odot$ is point-wise multiplication. This framework essentially attaches triggers to the sample piloted by a mask. Triggers can be either manually specified~\cite{BadNet,ReflectionECCV} or obtained through optimization~\cite{clean_fudan,Invisible2,Invisible_Zhong}. However, in either case, triggers is often universal for different samples and highly perceptible to humans. Previous works~\cite{spectral_defense,clean_fudan} have also revealed the vulnerability when facing PCA-style defense method and inefficacy when attacking multiple target classes. 


To overcome above weakness, we design a novel solution with inspired ideas from generative models on point clouds~\cite{pointflow,foldingnet}. Specifically, we propose Poisoning MorphNet in a sample-adaptive manner, whose objective is described as below:
\begin{eqnarray}
\hat{x}_i = \M(x_i), \quad s.t. ~f(x_i) = y_i, \ \ f(\hat{x}_i) = t, \label{eqn:v1}
\end{eqnarray}
where $\M$ abstracts our generative MorphNet and $f$ denotes the backdoor victim classifier. Let $y_i$ and $t$ be the original class and target class of the sample, respectively. MorphNet reads a benign sample $x_i$ and returns a poisoned $\hat{x}_i$. Essentially, we optimize $\M(\cdot)$ such that $\hat{x}_i$ is visually similar to $x_i$, and stealthily conceals the trigger required for the attack (since $f(\hat{x}_i) = t$ is also enforced).

Figure~\ref{figure:structure} depicts the neural design of Poisoning MorphNet. It takes three considerations into account, which progressively refines and instantiates the idea in Eqn.~\ref{eqn:v1}:

\textbf{\rom{1}: two-branch structure for spectral separation.} Critically, MorphNet seeks to simultaneously optimize two objectives which seemingly contradict each other, namely high visual similarity (thus $\| x_i - \hat{x}_i\|$ should be small) and strong trigger embedding (thus $\| x_i - \hat{x}_i\|$ should be sufficiently large to satisfy $f(\hat{x}_i) = t$). We resolve the contradiction via an implicitly spectral analysis method. Our main insight is, point clouds can be spectrally decomposed into low-frequency and high-frequency components. The former scaffolds the key geometric structure, and the latter depicts more local fine-grained details. Such a spectral decomposition sheds light on achieving both goals of visual similarity (through maximally preserving low-frequency information in $x_i$) and trigger injection (\ie, poisoning $x_i$ by morphing the high-frequency signal). Formally, we can refine Eqn.~\ref{eqn:v1} via:
\begin{equation}
    \hat{x}_i = \M(x_i) = \T_2 ( concat[\T_1(x_i) , \P(\T_1(x_i))] ), \label{eqn:v2}
\end{equation}
where $\T_1$ and $\T_2$ are both multi-layer perceptron (MLP) functions that do the job of feature processing. $concat()$ is the operator of vanilla feature concatenation. As shown in Figure~\ref{figure:structure}, the module $\T_1$ first squeezes the key information of $x_i$. The network then diverges into two branches: a residual branch that conveys $\T_1(x_i)$ untouched, and a poisoning branch $\P(\T_1(x_i))$ is responsible for modifying $x_i$ and hiding triggers in it. Inspired by AutoEncoder~\cite{AE,Achlig3d} based generative model, we implement $\P$ as following:
\begin{eqnarray}
z = Encoder(d),~~\P(d) = Decoder(z), \label{eqn:p1}
\end{eqnarray}
where the module names $Encoder()$ and $Decoder()$ tell their functionalities. 

For the encoder, we design it as a combination of MLP and graph local max-pooling layer, following the successful practice in mainstream graph-based point cloud models~\cite{Achlig3d,DGCNN}. The architecture of the decoder part is a significantly-improved version of FoldingNet~\cite{foldingnet}. To be specific, it repeats $z$ to be consistent with the number of points, and each point is concatenated with coordinates uniformly sampled from a fixed grid. The original implementation in FoldingNet samples from a 2-D rectangle, which unreasonably constrains the number of points in the output to some squared integer. In MorphNet, we instead sample coordinates from a 3-D sphere, which makes the sampling process more flexible and elegantly align the dimensions of input / output points. Different from the previous works, this MorphNet-based sample-generating method is sample-variant, since the poisoning branch adaptively find the best positions to hide the trigger information. More network details are deferred to the supplemental material.

\textbf{\rom{2}: label encoding for conditional morphing.} For multi-target backdoor attack, in order to reduce the cost of repetitive selection of multiple triggers, we desire the flexibility of poisoning $x_i$ with different target class $t$.  MorphNet in Eqn.~\ref{eqn:v2} is further refined by incorporating target classes:
\begin{eqnarray}
\hat{x}^t_i = \M(x_i,t), \ \ s.t.~~f(x_i) = y_i, ~f(\hat{x}^t_i) = t, \label{eqn:v3}
\end{eqnarray}
where the new $\M$ generates a poisoned sample $\hat{x}^t$ conditioned on an arbitrary target class $t$. The superscript $t$ in $\hat{x}^t$ emphasizes its dependence on $t$. This enables the attack to multiple target classes using a single model.


Accordingly, the poisoning branch previously defined in Eqn.~\ref{eqn:p1} is converted to be conditional AutoEncoder~\cite{CVAE}. In practice, we concatenate the compressed representation $z$ obtained from encoder with a one-hot code $H_t$ (all-zero except for the $t$-th element) that encodes the target class $t$:
\begin{align}
z = Encoder(d),~\P(d) = Decoder(concat(z, H_t)). \label{eqn:z3}
\end{align}



\textbf{\rom{3}: progressive refinement using stacked blocks}. Previous clean-label attack methods~\cite{clean_fudan,ReflectionECCV,cleanlabel_turner} all sacrifice trigger's low visibility for high attack success rate. It is non-trivial to generate poisoned sample which is both visually similar to the clean version and easily recognized by the victim model. One natural remedy is to adopt more complex trigger-embedding module. To further promote MorphNet, we unroll $\M$ multiple times and stack them:
\begin{align}
\hat{x}_{1_i} &= \M_1(x_i,t), \nonumber\\
\hat{x}_{2_i} &= \M_2(\hat{x}_{1_i} + x_i,t),\nonumber\\
&......\nonumber\\
\hat{x}^t_i = \hat{x}_{n_i} &= \M_n(\hat{x}_{(n-1)_i} + x_i,t),
\end{align}
where $\M_k$ is the $k$-th instance of $\M$, described as in Eqns.~\ref{eqn:v3} and~\ref{eqn:z3}. $\hat{x}_{n_i}$ is the $n$-th intermediate output. Notably, there is also a residual link in each layer that feeds the benign input $x_i$ to every layer, which is designed to preserve the main structure of $x_i$. The unrolled $\M$ gradually improves the performance with diminishing gains, as later demonstrated by our experiments.


\subsection{Threat model}

\textbf{Basic Poisoning MorphNet}. We first train MorphNet using a cleanly-trained model $f^{clean}$ as simulated target model. To enable that MorphNet could attack multi-targets, we use an adversarial classification loss, where all categories are traversed as the target class, regardless of the ground-truth class of the sample:
\begin{small}
\begin{align}
    \mathcal{L}_{cls} =& - \sum_{t \in \mathcal{C}}\sum_{x_i \in \mathcal{D}_{train}} log(f^{clean}_{t}(\hat{x}^t_i)) \nonumber \\
  =& - \sum_{t \in \mathcal{C}}\sum_{x_i \in \mathcal{D}_{train}} log(f^{clean}_{t}(\M(x_i,t))),
\end{align}
\end{small}
where $f^{clean}_t$ is the $t$-th softmaxed score given by $f^{clean}$, indicating the classification score for class $t$. Besides, in order to keep the similarity between the generated sample and the original sample, an reconstruction loss $\mathcal{L}_{rec}$ is also used:
\begin{equation}
    \mathcal{L}_{rec} = \sum_{t \in \mathcal{C}}\sum_{x_i \in \mathcal{D}_{train}} \sum_{k = 1,2...,n} l_{chamfer}(x_i, \hat{x}^t_{k_i}),
\end{equation}
where we aggregate the Chamfer distance~\cite{chamfer} between the benign sample and all intermediate output, enforcing a progressive refinement:
\begin{eqnarray}
\label{eq:chamfer}
    &&l_{chamfer}(x_i, \hat{x}^t_i) \nonumber \\
    &=& \sum_{p_1 \in x_i} \mathop{min}_{p_2 \in \hat{x}^t_i} ||p_1 - p_2 ||^2 + \sum_{p_2 \in \hat{x}^t_i} \mathop{min}_{p_1 \in x_i} ||p_1 - p_2 ||^2.\nonumber
\end{eqnarray}
The loss of basic MorphNet weighs above two losses $\mathcal{L} =  \mathcal{L}_{cls} + \lambda \mathcal{L}_{rec}$, with $\lambda$ as the weighting hyper-parameter.


\textbf{MorphNet$^\ast$: enhanced model with denoising loss}. Denoising~\cite{SOR,lg_gan} is known to be particularly effective to resist various attacks on point clouds. It eliminates a large body of outliers in a sample, where adversarial signals or triggers conceal. To make MorphNet more resistant to such defense methods, we devise a third denoising loss which suppresses the outliers in poisoned samples generated by the MorphNet. 


To be specific, for each point $p_i$ in point cloud $x$, we first find its $k$-nearest neighbors ($k=3$ in practice), forming an index set $\mathcal{N}_i$. We use the notation $D(p_i)$ to denote the average distance of $p_i$ to its $k$ neighbors, namely $D(p_i) = 1/k \cdot \sum_{p_j \in \mathcal{N}_i} \| p_i - p_j \|$. Afterwards, we pick up $m$ (set to 30 in the experiments) points with largest $D(\cdot)$, obtaining an index set $\mathcal{N}_x$. Those points are supposed to most possible ``outliers". We abuse the notation $D$ to measure the degree of outliers in $x$, as $D(x) = 1/m \cdot \sum_{p_i \in \mathcal{N}_x} D(p_i)$. The denoising loss is defined by the aggregation over all generated poisoned samples:
\begin{align}
   \mathcal{L}_{den} = \sum_{t \in \mathcal{C}}\sum_{x_i \in \mathcal{D}_{train}}  D(\hat{x}^t_i),
\end{align}
which intuitively encourages fewer wild points for countering the denoising operation. We denote the denoising-enhanced version as MorphNet$^\ast$, whose objective is $\mathcal{L} =  \mathcal{L}_{cls} + \lambda \mathcal{L}_{rec} + \theta \mathcal{L}_{den}$, with $\lambda$ and $\theta$ as hyper-parameters.




\textbf{Attack with poisoned samples}. Under multi-target clean-label setting, for each class $t \in \mathcal{C}$, we take $\alpha \%$ ($\alpha \%$ is the poison rate) clean samples and morph them into poisoned samples conditioned on $t$. After data generation, we inject the poisoned sample from all classes into the training set to train the victim model. The victim model can recognize triggers once converged.  At inference time, one can attack the victim model by generating a poisoned sample conditioned on any target class.


\section{Evaluations}

\subsection{Data Preparation and Evaluation Protocols}

\textbf{Datasets and recognition model.} We adopt two widely-used point cloud datasets: ModelNet40~\cite{modelnet} and ShapeNetcorev2~\cite{shapenet}. The former contains 12,311 pre-aligned samples from 40 categories, split into 9,843 for training and 2,468 for testing. The latter contains 51,127 pre-aligned samples from 55 categories, with 35,708 for training and 10,261 shapes for testing. Following~\cite{pointnet}, we uniformly sample 2,048 points from each sample on both datasets. For recognition model, we use PointNet~\cite{pointnet} as our victim model, and further explore the transferability between different recognition models, \eg, DGCNN~\cite{DGCNN} and PointNet++~\cite{pointnet++}.


\textbf{Baselines.} Since MorphNet is the first work of its kind, we tailor two state-of-the-art clean-label backdoor attack methods originally for images and videos into point cloud as our baselines: Turner et al.~\cite{cleanlabel_turner} and Zhao et al.~\cite{clean_fudan}. For the adversarial perturbation used by~\cite{cleanlabel_turner,clean_fudan}, we use PGD to perturb the benign samples same to~\cite{clean_fudan}. For the static trigger used in~\cite{cleanlabel_turner}, we randomly sample 20 points from the point cloud, and rearrange them into a straight line emitting from the origin to certain angle. For the adversarial trigger used in~\cite{clean_fudan}, same operation as in~\cite{clean_fudan} are adopted to obtain the optimized trigger. We initialize the trigger with 20 randomly selected points from the sample, attach the trigger on the non-target sample and optimize it, such that the benign model would misclassify it as target class. 5 classes are randomly drawn from ModelNet40 as the target classes. 

We would emphasize that both~\cite{cleanlabel_turner,clean_fudan} are designed for single-target attack and need to iterate the whole optimization for each target class. In contrary, our proposed MorphNet works in a multi-tagret attack setting, attacking all classes in a single poisoning process. For a fair comparison with Zhao et al. \cite{clean_fudan}, we also add a denoising loss in its trigger optimization process to enhance its performance on defended model.

\textbf{Attack settings}. MorphNet is implemented in PyTorch~\cite{pytorch} and trained using Adam with an initial learning rate of 0.0001 and batchsize of 16 for 200 epochs. $\lambda$ and $\theta$ are set as 0.05 and 0.02 by default. We train the target model using SGD with an initial learning rate of 0.1 and batchsize of 32 for 250 epochs.


To evaluate the attack performance of certain target class, we use MorphNet to generate poison samples conditioned this target class for all test data. Then Attack Success Rate (ASR) is obtained as the proportion of samples misclassified as the target class by the victim model. mASR is further calculated by averaging the ASR of all classes. Besides, we also test the model that uses Statistical Outlier Removal (SOR) denoising defense~\cite{SOR} for removing outliers before feeding the sample to the victim model. Hyper-parameters of SOR are set following~\cite{SOR}. The attack results are denoted as mASR-D, which `D' implies the \emph{denoising} defense. Surprisingly, the accuracy of the poisoned model on clean test set has no obvious decline compared with the benign model in most cases. For saving space, we defer the detailed report of the former in the supplemental.



\subsection{Experimental Results}

Main results on two datasets are shown in Table~\ref{tab:main}. Enhanced by denoising loss, MorphNet$^\ast$ achieves better attack performance on defended models ($5.6 \rightarrow 17.1$ and $4.8 \rightarrow 15.1$, respectively). As the cost, the denoising loss limits the generation of wild points, which sacrifices the attack performance on undefended model to some extent ($82.4 \rightarrow 62.8$ and $76.7 \rightarrow 57.2$, respectively). With fewer wild points, the reconstruction performance of MorphNet$^\ast$ is slightly better than MorphNet. 



\begin{table}[t]
\centering
\begin{footnotesize}
\begin{tabular}{c c c  |c}
\hline
Dataset & Model &mASR/mASR-D& Chamfer Dist. \\
\hline
\multirow{2}*{ModelNet40}&MorphNet& \textbf{82.4} / 5.6& 3.81\\
&MorphNet$^\ast$& 62.8 / \textbf{17.7}& 3.78\\
\hline
\multirow{2}*{ShapeNetcorev2}&MorphNet &\textbf{76.7} / 4.8& 3.56\\
&MorphNet$^\ast$& 57.2 / \textbf{15.1} & 3.53\\
\hline
\end{tabular}
\end{footnotesize}
\caption{\small Attack performance on ModelNet40 and ShapeNetcorev2 in terms of mASR/mASR-D (\%) and reconstruction loss evaluated by Chamfer Distance.}
\label{tab:main}
\vspace{-0.1in}
\end{table}

\begin{table*}[t]
\centering
\begin{footnotesize}
\begin{tabular}{c |c c c c c|c|c}
\hline

Method &\multicolumn{5}{c|}{ASR/ASR-D} &mASR/mASR-D $\uparrow$ & Chamfer Distance $\downarrow$\\
 & airplane & bottle& guitar&bed&monitor& & \\
\hline
Turner et al. \cite{cleanlabel_turner}& 7.4 / 3.4&15.8 / 2.6& 8.1 / 4.1 &11.4 / 2.4&16.2 / 3.5&11.7 / 3.2 &7.12 \\
Zhao et al. \cite{clean_fudan}&13.4 / 4.3 &21.6 / 2.3&19.4 / 3.8&20.4 / 3.7&17.1 / 2.4& 18.3 / 3.3&8.06\\
Zhao et al. \cite{clean_fudan} with $\mathcal{L}_{den}$ &9.1 / 8.3 &18.5 / 5.3&14.5 / 4.8&15.9 / 4.2&14.0 / 5.1& 14.4 / 5.6&7.91\\
MorphNet (ours) &\textbf{86.3} / 9.6 & \textbf{58.6} / 4.8& \textbf{85.1} / 4.3& \textbf{70.5} / 3.7& \textbf{77.1} / 4.5&\textbf{75.5} / 5.4& 3.81\\
MorphNet$^\ast$(ours)&74.1 / \textbf{20.8} & 33.3 / \textbf{8.1} & 73.9 / \textbf{18.8}& 55.0 / \textbf{11.6}& 71.2 / \textbf{15.5}&61.5 /\textbf{15.0} & \textbf{3.78}\\
\hline
\end{tabular}
\end{footnotesize}
\caption{\small Attack performance compared with baselines on ModelNet40 in terms of ASR (\%) etc.}
\label{tab:baseline}
\vspace{-0.1in}
\end{table*}

\begin{figure}[t]
\centering
\includegraphics[width=\linewidth]{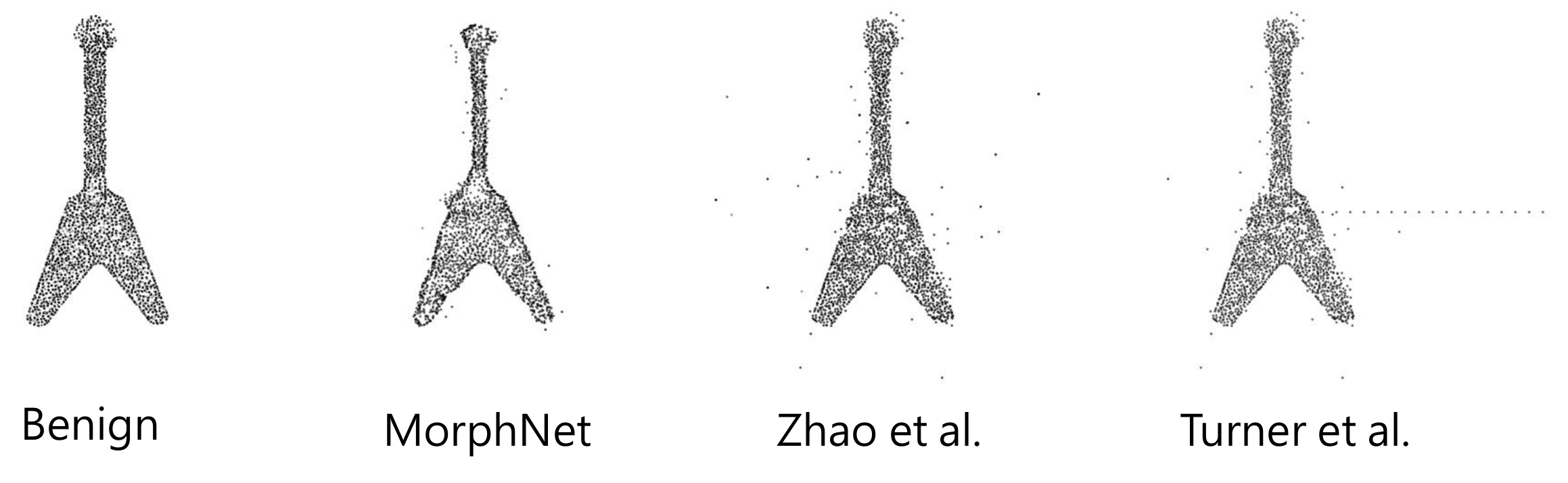}
\caption{\small Qualitative comparison of poisoned samples generated by MorphNet and two baselines~\cite{cleanlabel_turner,clean_fudan}. Better viewing when enlarged.}
\label{figure:qualitative results baseline}
\vspace{-0.1in}
\end{figure}




Comparisons with baselines are shown in Table~\ref{tab:baseline}. When attacking undefended model, our method significantly surpasses both baselines. Enhanced with the denoising loss, the attack performance of Zhao et al. \cite{clean_fudan} on defended model demonstrates clear improvement (3.3 $\rightarrow$ 5.6), and the attack performance on undefended model slightly decreases (18.3 $\rightarrow$ 14.4), which is consistent to the observation in Table~\ref{tab:main}. Under all evaluation metrics, our proposed MorphNet$^\ast$ clearly outruns the baselines when attacking the defended models. Regarding the trigger visibility, the mean Chamfer distance between our generated sample and benign sample is also consistently smaller than baseline's. 

Visualization of some poisoned samples is found in Figure~\ref{figure:qualitative results baseline}. As seen, the fixed trigger proposed by~\cite{cleanlabel_turner} can be easily distinguished from the benign sample.The optimized trigger in~\cite{clean_fudan} introduces many outliers. Our generated sample has fewer outliers, and keeps the same salient structure to the benign sample while achieving a good attack performance.



\subsection{Ablation Study}

Due to the space limit, we here only report the attack results of MorphNet on undefended models. Full results on defended model and MorphNet$^\ast$ can be found in supplementary materials.


\textbf{Structure of MorphNet.} We first conduct ablation study to investigate the network designs of MorphNet in terms of attack performance and reconstruction capacity. In specific, we choose a base model FoldingNet~\cite{foldingnet} with similar complexity to MorphNet, and tailor it for generating samples according to the condition class. The results are found in Table~\ref{tab:structure}. Compared with the more generic FoldingNet, the 1-stacked MorphNet could significantly improve the reconstruction ability (6.31 \emph{v.s.} 3.68), which supposedly attributes to the structure-preserving main branch in MorphNet. When more stacks are used, the attack performance notably increases while the reconstruction loss still stays at a low level. Using even more stacks will have diminishing returns. Therefore, we choose 2 as our best stacking choice. To sum up, our stacked MorphNet can achieve better attack performance compared with FoldingNet (82.4 \textit{vs.} 81.3), meanwhile significantly improving the reconstruction capacity.

\begin{table}[t]
\centering
\begin{footnotesize}
\begin{tabular}{c c c c c}
\hline
Model & mASR $\uparrow$ &  Chamfer Distance $\downarrow$\\
\hline
FoldingNet \cite{foldingnet}& 81.3 & 6.31\\
\hline
1-stacked MorphNet& 57.4 & \textbf{3.68}\\
2-stacked MorphNet&\textbf{82.4} &  3.81\\
3-stacked MorphNet& 79.9&  4.02\\
\hline
\end{tabular}
\end{footnotesize}
\vspace{0.1in}
\caption{\small Ablation studies of model structures.}
\label{tab:structure}
\vspace{-0.1in}
\end{table}

\begin{figure}[t]
\centering
\includegraphics[width = \linewidth]{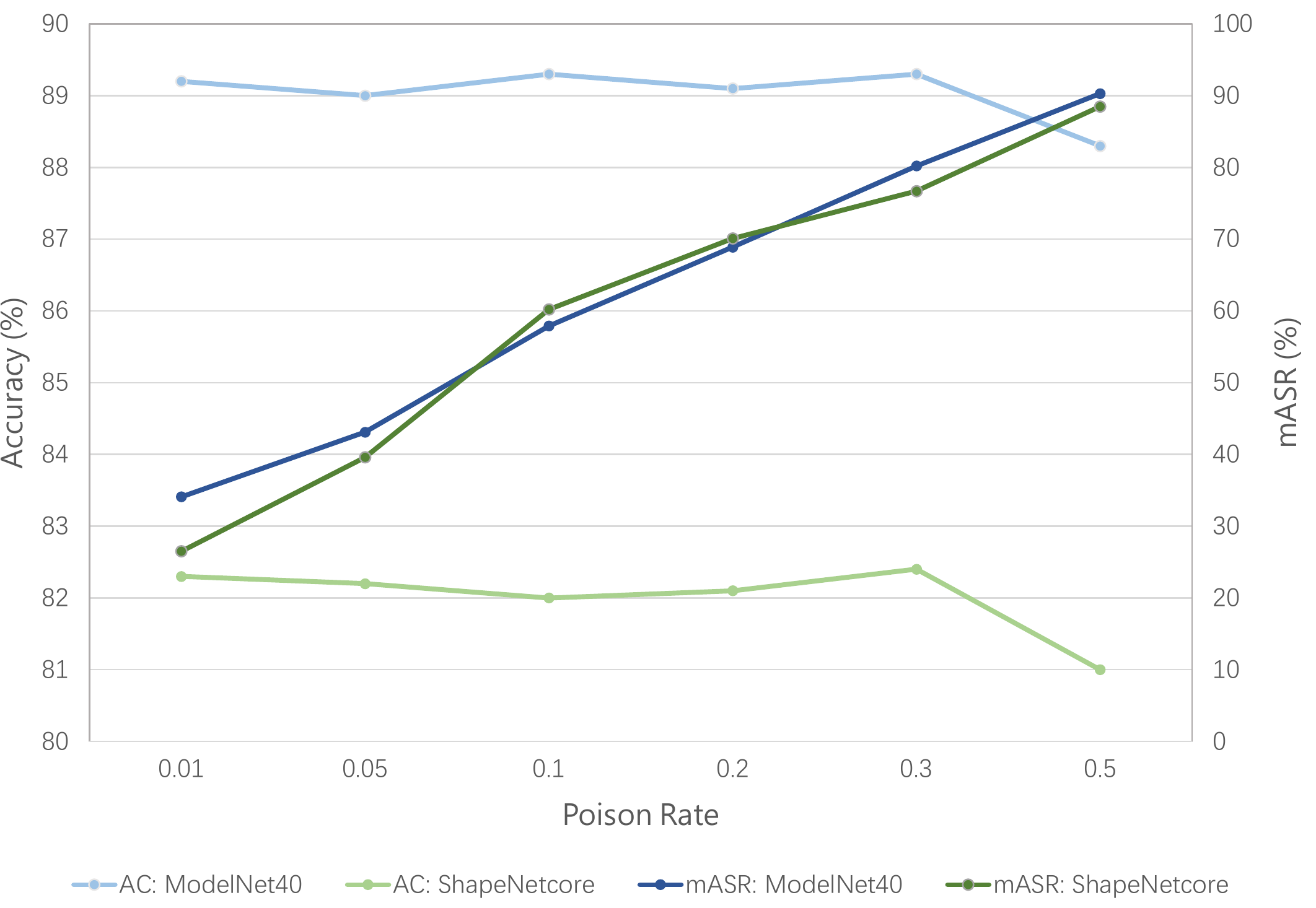}
\caption{\small Attack performance \textit{vs.} poison rate.}
\label{figure:poison_rate}
\vspace{-0.1in}
\end{figure}

\textbf{Attack performance \textit{vs.} poison rate.} We further conduct experiments to investigate the effect of poison rate on the attack performance. Results are found in Figure~\ref{figure:poison_rate}. When the poison rate increases, the attack performance increases accordingly. As seen, a poison rate below 30\% does not affect much the accuracy of the model on clean testing set. However, when the poison rate reaches 50\%, the accuracy of the model will decrease by a non-trivial extent (about 1\%). Nevertheless, with the poison rate as 30\%, our method can still achieve a very impressive attack performance (over 80\% mean success rate). Therefore, as long as the poisoning rate is reasonably controlled, our method can achieve a good attack performance without much sacrifice of the clean accuracy of the model.

\begin{figure}[t]
\centering
\includegraphics[width = \linewidth]{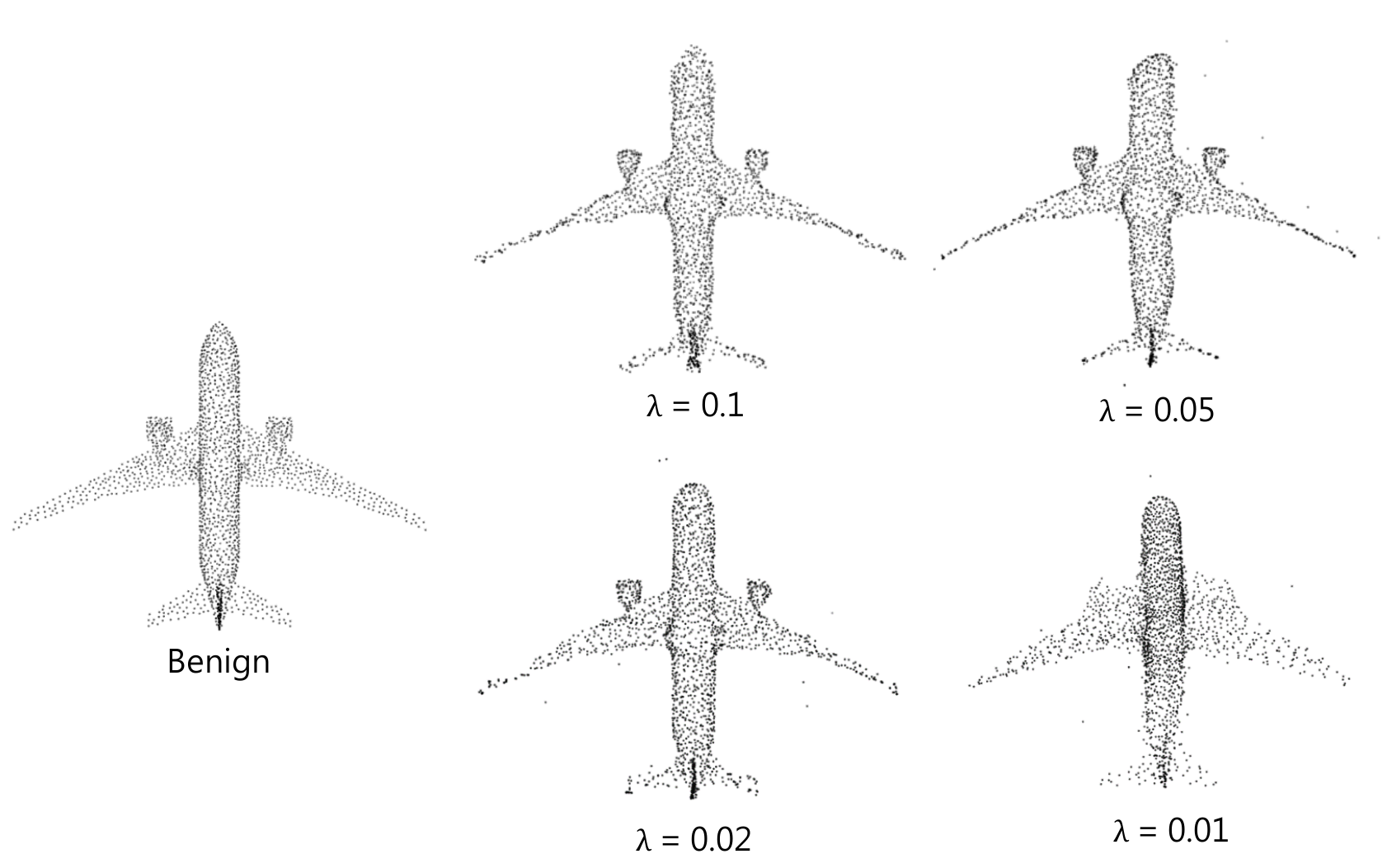}
\caption{\small Qualitative results with different $\lambda$s in MorphNet. Better viewing when enlarged.}
\label{figure:Qualitative_rec_loss}
\vspace{-0.1in}
\end{figure}

\textbf{Attack performance \textit{vs.} reconstruction loss.} We also check the effect of hyper-parameter $\lambda$ that balances the attack performance and reconstruction capacity. Results are shown in Table~\ref{tab:rec_loss}. Intuitively, a larger value of $\lambda$ is more biased towards visual reconstruction, which will improve the quality of the generated samples yet reduce the attack performance. Qualitative results at different $\lambda$s are shown in Figure~\ref{figure:Qualitative_rec_loss}. In our experiments, we set $\lambda$ as 0.02 to get a good trade-off between reconstruction and attack.

\begin{table}[t]
\centering
\begin{footnotesize}
\begin{tabular}{c c c c c}
\hline
$\lambda$ & 0.1 & 0.05& 0.02 & 0.01 \\
\hline
mASR& 40.6 & 57.4 & 82.4 & 83.1\\

Chamfer Distance& 2.12 & 3.28&3.81&6.37\\

\hline
\end{tabular}
\end{footnotesize}
\vspace{0.1in}
\caption{\small Attack performance \textit{vs.} reconstruction loss with different $\lambda$s in the proposed MorphNet. }
\label{tab:rec_loss}
\vspace{-0.1in}
\end{table}

\begin{figure*}[t]
\centering
\includegraphics[width = 0.85\linewidth]{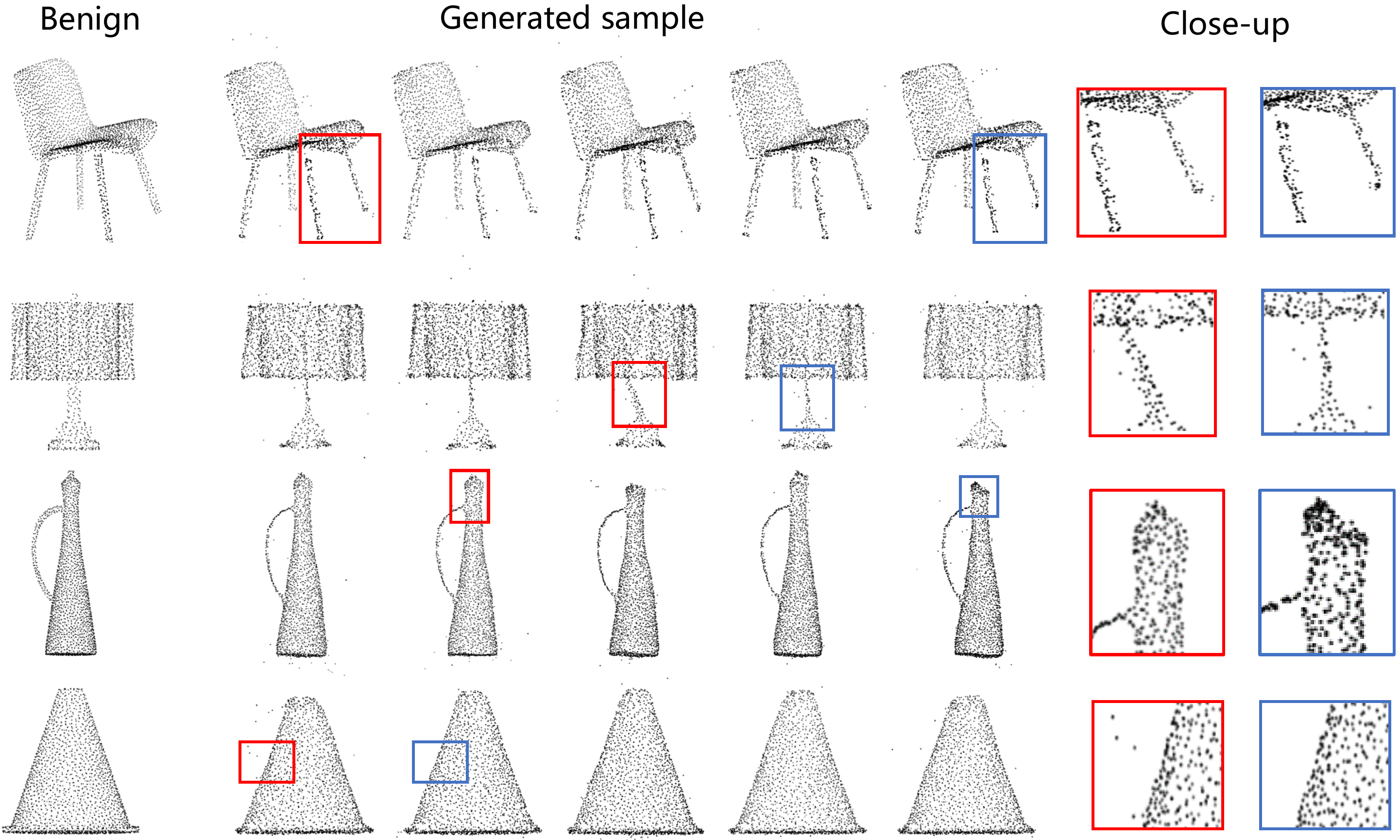}
\caption{\small Qualitative results from ModelNet40.}
\label{figure:qualitative results}
\vspace{-0.05in}
\end{figure*}

\subsection{Transferability}

In practice, the attacker may not be knowledgable about the specific structure of target models. Equivalently stated, the model used for training MorphNet may be different from the target models. In previous adversarial attack methods on point cloud~\cite{lg_gan,xiang3d}, such a structural difference can lead to attack failure, implying a low transferability.

To validate the transferability of our generated samples, we choose two other representative models as our target models, including DGCNN~\cite{DGCNN} and PointNet++~\cite{pointnet++}, which differ from PointNet for their neural designs. We firstly train MorphNet with a benign PointNet, and inject the poison samples generated by the trained MorphNet into the training set of different recognition models. The poison rate is set as 30\%. Table~\ref{tab:transfer} shows the attack performance and accuracy change ($\Delta$ accuracy) on benign test data. When transferred to a different model, the attack performance significantly drops yet still stay on a reasonable level. This shows a good transferability of our method. In addition, the test accuracy on the benign data has no obvious decline compared with the benign model (sometimes even higher than the benign model), which shows that our generated poisoned sample has very similar distribution to the benign samples.

\begin{table}[t]
\centering
\begin{footnotesize}
\begin{tabular}{c c c c }
\hline
Model& Recognition Model & mASR/mASR-D &  $\Delta$ accuracy \\
\hline
\multirow{3}*{MorphNet} &PointNet \cite{pointnet}& 82.4 / 5.6&  $\pm$ 0.1\\
&DGCNN \cite{DGCNN}& 42.4 / 4.9 & $\pm$ 0.2\\
&PointNet++ \cite{pointnet++}&49.9 / 5.2 & $\pm$ 0.1 \\
\hline
\multirow{3}*{MorphNet$^\ast$} &PointNet \cite{pointnet}& 62.8 / 17.7 & $\pm$ 0.1\\
&DGCNN \cite{DGCNN}& 32.5 / 12.5 & $\pm$ 0.1\\
&PointNet++ \cite{pointnet++}&39.1 / 14.6 & $\pm$ 0.2 \\
\hline
\end{tabular}
\end{footnotesize}
\vspace{-0.1in}
\caption{\small Transfer results on DGCNN and PointNet++ in terms of mASR/mASR-D (\%) and $\Delta$ accuracy (\%).}
\label{tab:transfer}
\end{table}

\subsection{Qualitative Results}

Figure~\ref{figure:qualitative results} presents some qualitative results of the generated poison samples by MorphNet based on several samples from ModelNet40 and their partial close-ups. The leftmost column contains the benign images. The 5 columns in the middle are the generated samples conditioned on different classes. To give a better view on the local details, we provide close-ups in the 2 rightmost columns, which contain 2 local areas with largest differences in the same row. As can be observed, when conditioned on different classes, the generated samples from the same benign one would vary in local details. In a majority of cases, the generated samples preserve the same salient structure as the benign sample, with only a few outliers produced. Rarely there are wild differences with respect to the benigh that can be noticed (\eg, the pillar of lamp in the second row and third column is bent). More qualitative results from MorphNet$^\ast$ can be found in supplementary material.

\subsection{Resistance to Defense Methods}

\textbf{Data augmentation.} In the point cloud recognition task, data enhancement is a common way to improve the model performance, including random rotation, random jitter, etc. It is also popularly used to resist attacks, because it may change or destroy the adversarial signals or triggers. In order to verify whether data augmentation can resist MorphNet, we conduct data augmentation in the training phase of the target model. Following~\cite{foldingnet}, we apply a random rotation that is one of the 24 axis-aligned in right-handed system. For random translation, we randomly scale the point cloud and jitter the position of each points by random noises with zero mean. Results are shown in the left of Table~\ref{table:pac_aug}. It is shown that, data augmentation can reduce the attack performance, among which the random rotation is more resistant and can reduce the attack success rate by roughly 20 points. However, the attack performance is still at a relatively high level, \eg, 53.9\%, which shows that data augmentation does not provide a good defense against our attack method.



\begin{table}[t]
\begin{floatrow}
\begin{footnotesize}

\begin{tabular}{c c }
\hline
Augmentation & mASR  \\
\hline
None & 82.4\\
Rotation & 58.1 \\
Translation & 73.3 \\
Rotation \& Translation & 53.9 \\
\hline
\end{tabular}
\hspace{0.1in}
\begin{tabular}{c c }
\hline
Poison rate & 30 \\
\hline
mean proportion & 38.03 \\
min proportion & 28.9 \\
max proportion & 46.1 \\
\hline
\end{tabular}

\end{footnotesize}
\vspace{0.1in}

\end{floatrow}
\caption{\small Resistance to data augmentation (left) and spectral signature detection (right) on ModelNet40.}
\label{table:pac_aug}
\vspace{-0.2in}
\end{table}

\textbf{Spectral signature detection.} Tran et al.~\cite{spectral_defense} proposed Spectral Signatures to filter out the poisoned samples from the training set. We conduct experiments to test whether our generated sample can be detected by this defense method. In detail, given a certain class in the injected training set with a poison rate of 30\%, we calculate the cosine similarity between a feature and the top right singular vector of the feature matrix for every sample. Then, we sort the samples according to the cosine similarity, and take the top 50\% samples as the candidate poison samples. According to~\cite{spectral_defense}, poisoned sample with the fixed trigger shall have higher similarity score, so it will also appear in the candidate poison sample. We calculate the proportion of poisoned data in the candidate poison samples and take the average of all classes. It is found that the average proportion is only slightly higher (38.03\%) than the basline score (30\%), as shown in the right of Table~\ref{table:pac_aug}. Compared with the results reported in~\cite{clean_fudan} where around 93.3\% samples in the candidate set are poisoned sample, this result shows that the defense method does not detect our generated samples well. The main reason is supposed to be that our ``trigger" is sample-adaptive and leaves no obvious principal components in the feature matrix.

\begin{figure}[t]
\centering
\includegraphics[width = 0.85\linewidth]{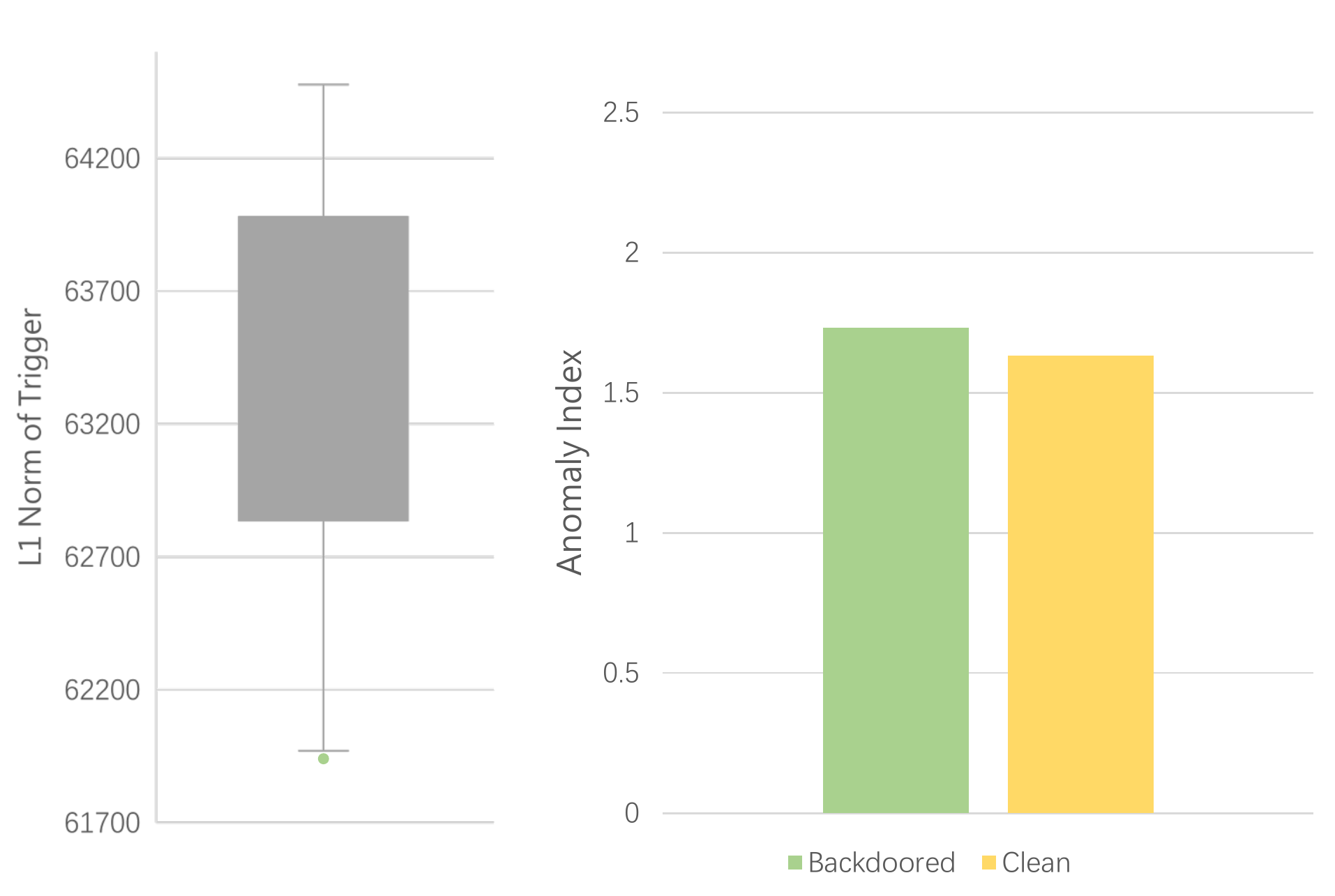}
\caption{\small Neural Cleanse results. Left: L1-norm of triggers for target class and non-target classes. We plot min/max, 25/75 quartile and median value of L1 norm for non-target classes and the target class are represented with dot. Right: Anomaly Index of backdoored model and clean model. A higher anomaly index represents a more significant minimum outlier, which means that the model is more likely to be infected. A anomaly index $>$ 2 is considered as an infected model.}
\label{figure:neural_cleanse}
\vspace{-0.1in}
\end{figure}

\textbf{Neural cleanse.} The work in~\cite{neural_clean} aims to detect whether a model has been hidden with triggers. It assumes that the perturbation required for the infected model to misclassify samples into a target class is usually smaller than other classes. Thus, with outlier detection method like median absolute deviation (MAD), Neural Cleans can determine whether a model has been infected or not, if a certain class requires significantly smaller perturbation than other categories. Our attack method could attack all classes at the same time, and the perturbation required by all classes would stay at a same level. Therefore, Neural Cleanse is actually not applicable to our method. Nevertheless, to get a better understanding of our attack model, we simulate a scenario where only one target class is attacked, by only injecting poisoned samples for one target class, as done in previous clean-label backdoor attack methods. The results are shown in Figure~\ref{figure:neural_cleanse}, where our backdoored anomaly index is below 2, which means that Neural Cleanse fails to detect our attack. We suppose this is mainly because our sample-adaptive poisoned sample makes the defense method unable to find a suitable universal perturbation to fit our trigger.



\section{Conclusion}

We propose a novel model for attacking point clouds in a clean-label setting, dubbed as Poisoning MorphNet, which is the first work on clean-label backdoor attack for point cloud. We conduct comprehensive experiments to fully validate our method. The results show the effectiveness and excellent concealment of our method.

{\small
\bibliographystyle{ieee_fullname}
\nocite{*}
\bibliography{egbib}

\begin{thebibliography}{10}\itemsep=-1pt

\bibitem{Achlig3d}
Panos Achlioptas, Olga Diamanti, Ioannis Mitliagkas, and Leonidas~J. Guibas.
\newblock Representation learning and adversarial generation of 3d point
  clouds.
\newblock {\em CoRR}, abs/1707.02392, 2017.

\bibitem{adt_survey}
Naveed Akhtar and Ajmal~S. Mian.
\newblock Threat of adversarial attacks on deep learning in computer vision:
  {A} survey.
\newblock {\em {IEEE} Access}, 6:14410--14430, 2018.

\bibitem{AE}
Pierre Baldi.
\newblock Autoencoders, unsupervised learning, and deep architectures.
\newblock In {\em ICML}, pages 37--50, 2012.

\bibitem{CW}
Nicholas Carlini and David~A. Wagner.
\newblock Towards evaluating the robustness of neural networks.
\newblock In {\em 2017 {IEEE} Symposium on Security and Privacy, {SP} 2017, San
  Jose, CA, USA, May 22-26, 2017}, pages 39--57, 2017.

\bibitem{i3d}
Jo{\~{a}}o Carreira and Andrew Zisserman.
\newblock Quo vadis, action recognition? {A} new model and the kinetics
  dataset.
\newblock In {\em CVPR}, pages 4724--4733, 2017.

\bibitem{shapenet}
Angel~X. Chang, Thomas~A. Funkhouser, Leonidas~J. Guibas, Pat Hanrahan,
  Qi{-}Xing Huang, Zimo Li, Silvio Savarese, Manolis Savva, Shuran Song, Hao
  Su, Jianxiong Xiao, Li Yi, and Fisher Yu.
\newblock Shapenet: An information-rich 3d model repository.
\newblock {\em CoRR}, abs/1512.03012, 2015.

\bibitem{Invisible1}
Xinyun Chen, Chang Liu, Bo Li, Kimberly Lu, and Dawn Song.
\newblock Targeted backdoor attacks on deep learning systems using data
  poisoning.
\newblock {\em CoRR}, abs/1712.05526, 2017.

\bibitem{drive}
Xiaozhi Chen, Huimin Ma, Ji Wan, Bo Li, and Tian Xia.
\newblock Multi-view 3d object detection network for autonomous driving.
\newblock In {\em CVPR}, pages 6526--6534, 2017.

\bibitem{chamfer}
Haoqiang Fan, Hao Su, and Leonidas~J. Guibas.
\newblock A point set generation network for 3d object reconstruction from a
  single image.
\newblock In {\em CVPR}, pages 2463--2471. {IEEE} Computer Society, 2017.

\bibitem{survey3}
Yansong Gao, Bao~Gia Doan, Zhi Zhang, Siqi Ma, Jiliang Zhang, Anmin Fu, Surya
  Nepal, and Hyoungshick Kim.
\newblock Backdoor attacks and countermeasures on deep learning: {A}
  comprehensive review.
\newblock {\em CoRR}, abs/2007.10760, 2020.

\bibitem{FGSM}
Ian~J. Goodfellow, Jonathon Shlens, and Christian Szegedy.
\newblock Explaining and harnessing adversarial examples.
\newblock In {\em ICLR}, 2015.

\bibitem{BadNet}
Tianyu Gu, Brendan Dolan{-}Gavitt, and Siddharth Garg.
\newblock Badnets: Identifying vulnerabilities in the machine learning model
  supply chain.
\newblock {\em CoRR}, abs/1708.06733, 2017.

\bibitem{resnet}
Kaiming He, Xiangyu Zhang, Shaoqing Ren, and Jian Sun.
\newblock Deep residual learning for image recognition.
\newblock In {\em CVPR}, pages 770--778, 2016.

\bibitem{Invisible2}
Shaofeng Li, Benjamin Zi~Hao Zhao, Jiahao Yu, Minhui Xue, Dali Kaafar, and
  Haojin Zhu.
\newblock Invisible backdoor attacks against deep neural networks.
\newblock {\em CoRR}, abs/1909.02742, 2019.

\bibitem{survey1}
Yiming Li, Baoyuan Wu, Yong Jiang, Zhifeng Li, and Shu{-}Tao Xia.
\newblock Backdoor learning: {A} survey.
\newblock {\em CoRR}, abs/2007.08745, 2020.

\bibitem{liu3d}
Daniel Liu, Ronald Yu, and Hao Su.
\newblock Extending adversarial attacks and defenses to deep 3d point cloud
  classifiers.
\newblock In {\em ICIP}, pages 2279--2283, 2019.

\bibitem{ReflectionECCV}
Yunfei Liu, Xingjun Ma, James Bailey, and Feng Lu.
\newblock Reflection backdoor: {A} natural backdoor attack on deep neural
  networks.
\newblock {\em ECCV}, 2020.

\bibitem{survey2}
Yuntao Liu, Ankit Mondal, Abhishek Chakraborty, Michael Zuzak, Nina Jacobsen,
  Daniel Xing, and Ankur Srivastava.
\newblock A survey on neural trojans.
\newblock {\em {IACR} Cryptol. ePrint Arch.}, 2020:201, 2020.

\bibitem{voxnet}
Daniel Maturana and Sebastian~A. Scherer.
\newblock Voxnet: {A} 3d convolutional neural network for real-time object
  recognition.
\newblock In {\em IROS}, pages 922--928, 2015.

\bibitem{NIPS_aware}
Anh Nguyen and Anh Tran.
\newblock Input-aware dynamic backdoor attack.
\newblock {\em NeurIPS}, 2020.

\bibitem{pytorch}
Adam Paszke, Sam Gross, Francisco Massa, Adam Lerer, James Bradbury, Gregory
  Chanan, Trevor Killeen, Zeming Lin, Natalia Gimelshein, Luca Antiga, Alban
  Desmaison, Andreas K{\"{o}}pf, Edward Yang, Zachary DeVito, Martin Raison,
  Alykhan Tejani, Sasank Chilamkurthy, Benoit Steiner, Lu Fang, Junjie Bai, and
  Soumith Chintala.
\newblock Pytorch: An imperative style, high-performance deep learning library.
\newblock In {\em NeurIPS}, pages 8024--8035, 2019.

\bibitem{pointnet}
Charles~Ruizhongtai Qi, Hao Su, Kaichun Mo, and Leonidas~J. Guibas.
\newblock Pointnet: Deep learning on point sets for 3d classification and
  segmentation.
\newblock In {\em CVPR}, pages 77--85, 2017.

\bibitem{pointnet++}
Charles~Ruizhongtai Qi, Li Yi, Hao Su, and Leonidas~J. Guibas.
\newblock Pointnet++: Deep hierarchical feature learning on point sets in a
  metric space.
\newblock In {\em NeurIPS}, pages 5099--5108, 2017.

\bibitem{octnet}
Gernot Riegler, Ali~Osman Ulusoy, and Andreas Geiger.
\newblock Octnet: Learning deep 3d representations at high resolutions.
\newblock In {\em CVPR}, pages 6620--6629, 2017.

\bibitem{AAAI_hidden}
Aniruddha Saha, Akshayvarun Subramanya, and Hamed Pirsiavash.
\newblock Hidden trigger backdoor attacks.
\newblock In {\em AAAI}, 2020.

\bibitem{dynamic_salem}
Ahmed Salem, Rui Wen, Michael Backes, Shiqing Ma, and Yang Zhang.
\newblock Dynamic backdoor attacks against machine learning models.
\newblock {\em CoRR}, abs/2003.03675, 2020.

\bibitem{ECC}
Martin Simonovsky and Nikos Komodakis.
\newblock Dynamic edge-conditioned filters in convolutional neural networks on
  graphs.
\newblock In {\em CVPR}, pages 29--38, 2017.

\bibitem{CVAE}
Kihyuk Sohn, Honglak Lee, and Xinchen Yan.
\newblock Learning structured output representation using deep conditional
  generative models.
\newblock In {\em NeurIPS}, pages 3483--3491, 2015.

\bibitem{MVCNN}
Hang Su, Subhransu Maji, Evangelos Kalogerakis, and Erik~G. Learned{-}Miller.
\newblock Multi-view convolutional neural networks for 3d shape recognition.
\newblock In {\em ICCV}, pages 945--953, 2015.

\bibitem{adt1}
Christian Szegedy, Wojciech Zaremba, Ilya Sutskever, Joan Bruna, Dumitru Erhan,
  Ian~J. Goodfellow, and Rob Fergus.
\newblock Intriguing properties of neural networks.
\newblock In {\em ICLR}, 2014.

\bibitem{spectral_defense}
Brandon Tran, Jerry Li, and Aleksander Madry.
\newblock Spectral signatures in backdoor attacks.
\newblock In {\em NeurIPS}, pages 8011--8021, 2018.

\bibitem{cleanlabel_turner}
Alexander Turner, Dimitris Tsipras, and Aleksander Madry.
\newblock Label-consistent backdoor attacks.
\newblock {\em CoRR}, abs/1912.02771, 2019.

\bibitem{neural_clean}
Bolun Wang, Yuanshun Yao, Shawn Shan, Huiying Li, Bimal Viswanath, Haitao
  Zheng, and Ben~Y. Zhao.
\newblock Neural cleanse: Identifying and mitigating backdoor attacks in neural
  networks.
\newblock In {\em 2019 {IEEE} Symposium on Security and Privacy, {SP} 2019, San
  Francisco, CA, USA, May 19-23, 2019}, pages 707--723, 2019.

\bibitem{DGCNN}
Yue Wang, Yongbin Sun, Ziwei Liu, Sanjay~E. Sarma, Michael~M. Bronstein, and
  Justin~M. Solomon.
\newblock Dynamic graph {CNN} for learning on point clouds.
\newblock {\em {ACM} Trans. Graph.}, 38(5):146:1--146:12, 2019.

\bibitem{modelnet}
Zhirong Wu, Shuran Song, Aditya Khosla, Fisher Yu, Linguang Zhang, Xiaoou Tang,
  and Jianxiong Xiao.
\newblock 3d shapenets: {A} deep representation for volumetric shapes.
\newblock In {\em CVPR}, pages 1912--1920, 2015.

\bibitem{xiang3d}
Chong Xiang, Charles~R. Qi, and Bo Li.
\newblock Generating 3d adversarial point clouds.
\newblock In {\em CVPR}, pages 9136--9144, 2019.

\bibitem{adv_gan}
Chaowei Xiao, Bo Li, Jun{-}Yan Zhu, Warren He, Mingyan Liu, and Dawn Song.
\newblock Generating adversarial examples with adversarial networks.
\newblock In {\em IJCAI}, pages 3905--3911, 2018.

\bibitem{pointflow}
Guandao Yang, Xun Huang, Zekun Hao, Ming{-}Yu Liu, Serge~J. Belongie, and
  Bharath Hariharan.
\newblock Pointflow: 3d point cloud generation with continuous normalizing
  flows.
\newblock In {\em ICCV}, pages 4540--4549, 2019.

\bibitem{nibingbing3d}
Jiancheng Yang, Qiang Zhang, Rongyao Fang, Bingbing Ni, Jinxian Liu, and Qi
  Tian.
\newblock Adversarial attack and defense on point sets.
\newblock {\em CoRR}, abs/1902.10899, 2019.

\bibitem{foldingnet}
Yaoqing Yang, Chen Feng, Yiru Shen, and Dong Tian.
\newblock Foldingnet: Point cloud auto-encoder via deep grid deformation.
\newblock In {\em CVPR}, pages 206--215, 2018.

\bibitem{MV3}
Ze Yang and Liwei Wang.
\newblock Learning relationships for multi-view 3d object recognition.
\newblock In {\em ICCV}, pages 7504--7513, 2019.

\bibitem{MV2}
Tan Yu, Jingjing Meng, and Junsong Yuan.
\newblock Multi-view harmonized bilinear network for 3d object recognition.
\newblock In {\em CVPR}, pages 186--194, 2018.

\bibitem{clean_fudan}
Shihao Zhao, Xingjun Ma, Xiang Zheng, James Bailey, Jingjing Chen, and
  Yu{-}Gang Jiang.
\newblock Clean-label backdoor attacks on video recognition models.
\newblock In {\em CVPR}, 2020.

\bibitem{zheng3dsanliency}
Tianhang Zheng, Changyou Chen, Junsong Yuan, Bo Li, and Kui Ren.
\newblock Pointcloud saliency maps.
\newblock In {\em ICCV}, pages 1598--1606, 2019.

\bibitem{Invisible_Zhong}
Haoti Zhong, Cong Liao, Anna~Cinzia Squicciarini, Sencun Zhu, and David~J.
  Miller.
\newblock Backdoor embedding in convolutional neural network models via
  invisible perturbation.
\newblock In {\em {CODASPY} '20: Tenth {ACM} Conference on Data and Application
  Security and Privacy}, pages 97--108, 2020.

\bibitem{lg_gan}
Hang Zhou, Dongdong Chen, Jing Liao, Kejiang Chen, Xiaoyi Dong, Kunlin Liu,
  Weiming Zhang, Gang Hua, and Nenghai Yu.
\newblock {LG-GAN:} label guided adversarial network for flexible targeted
  attack of point cloud based deep networks.
\newblock In {\em CVPR}, pages 10353--10362, 2020.

\bibitem{SOR}
Hang Zhou, Kejiang Chen, Weiming Zhang, Han Fang, Wenbo Zhou, and Nenghai Yu.
\newblock Dup-net: Denoiser and upsampler network for 3d adversarial point
  clouds defense.
\newblock In {\em ICCV}, pages 1961--1970, 2019.

\end{thebibliography}
}

\end{document}